\documentclass[preprint,12pt,3p]{elsarticle}

\usepackage{amssymb}

\usepackage{hyperref}

\usepackage{natbib}

\journal{Arxiv} 

\begin{document}

\begin{frontmatter}

\title{Optical Illusions Images Dataset}

\author{Robert Max Williams}
\ead{robertmaxwilliams@gmail.com}
\ead{https://robertmaxwilliams.github.io/}

\author{Roman V. Yampolskiy}
\address{University of Louisville}

\begin{abstract}
Human vision is capable of performing many tasks not optimized for in its long
evolution. Reading text and  identifying artificial objects such as road
signs are both tasks that mammalian brains never encountered in the wild
but are very easy for us to perform. However, humans have discovered many
very specific tricks that cause us to misjudge color, size, alignment and
movement of what we are looking at. A better understanding of these
phenomenon could reveal insights into how human perception achieves these
feats. In this paper we present a dataset of 6725 illusion images gathered
from two websites, and a smaller dataset of 500 hand-picked images. We will
discuss the process of collecting this data, models trained on it, and the
work that needs to be done to make it of value to computer vision
researchers.
\end{abstract}


\end{frontmatter}


\section{Motivation}

Being able to understand and intentionally create illusions is currently only possible for humans. Being able to accurately recognize illusory patterns using a computer, and to generate novel illusion images, would represent a huge advancement in computer vision. Current systems are capable of predicting the effect of specific classes of illusions, such as 
color consistency illusions \cite{ROBINSON20071631} and 
length illusions \cite{Garibay2015} \cite{bertulis2001distortions}.
A reinforcement learning system learned to perceive color consistency illusion after training to predict color values where half of the image was covered in a tinted film \cite{6400580}, showing that perception of an illusion can emerge from the demands of seeing in a complicated world. It is also important to consider whether making a perceptual mistake similar to humans constitutes having a visual experience similar to humans \cite{DBLP:journals/corr/abs-1712-04020}.

Recent work on generative adversarial networks (GANs) \cite{karras2017progressive} has shown that high resolution images of faces can be created using a large dataset of 30,000 images. 
This size and quality of images is not available for optical illusions; naively applying their methods to this dataset does not have the same results, as discussed below. The number of static optical illusion images is in the low thousands, and the number of unique kinds of illusions is certainly very low, perhaps only a few dozen (for example, the Scintillating Grid illusion, Cafe Wall Illusion and other known categories). Creating a model capable of learning from such a small and limited dataset would represent a huge leap in generative models and understanding of human vision.


\begin{figure}
    \centering
    \includegraphics[scale=0.4]{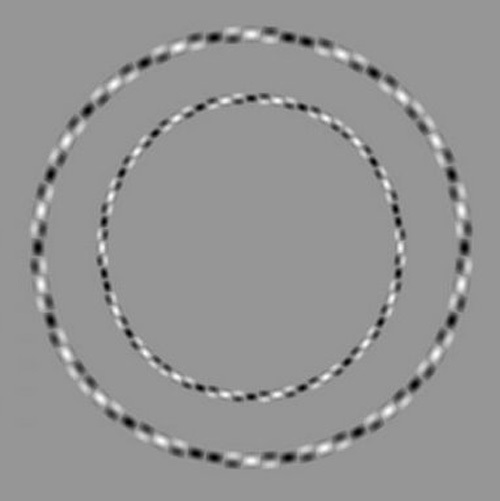}
    \caption{An illusion image from the dataset. The rings are circular and concentric, but the patterns and changes in contrast make them appear to be warped. Image credit to \url{viperlib.york.ac.uk}}
    \label{fig:whorls}
\end{figure}

\section{Related Works}

Research into biologically plausible models makes it possible to learn about visual phenomenon by doing experiments on proxies for the real human vision system. Elsayed et al. found that by selecting the right models, adversarial examples for these models were also effective on time-limited humans \cite{1802.08195}. The \textsc{Brain-Score} metric \cite{SchrimpfKubilius2018BrainScore} measures internal and behavioral similarity between computer and primate image recognition. As this metric is developed and models with higher scores are created, those model may be capable of experiencing more kinds of optical illusions otherwise only experienced by primates.

To our knowledge, no dataset of this kind has been created before.

\section{Data Collection}

\subsection{Image Sources}
Twelve different websites that collect and display optical illusions (such as one shown in Figure \ref{fig:whorls}) were considered for inclusion in the dataset. Most proved to be too small or not containing the right content. For instance, the site ``Visual Phenomena \& Optical Illusions" \cite{MichaelBach} contains many interesting and visually powerful demonstrations of optical illusions, but very few still images that by themselves contain a visual effect. In the end, ``Mighty Optical Illusions" \cite{MoIllusions} and ``ViperLib" \cite{ViperLib} proved to be the best sources of illusion images, both containing labeled images and containing almost exclusively static images.

\par
The ``Illusions of the Year" \cite{IllusionOfTheYear} contest also seemed to be a good source of images, but they only post the winning results publicly. Emails to the website owner requesting all of the submissions were not answered. 

\subsection{Data Collection Results}

We created a web scraper to go to each page of Mighty Optical Illusions and download the images on that page (source is available at \cite{OpticalIllusionDataset}). In total, 6436 images obtained, along with their metadata such as categories and page titles. ViperLib was scraped in a similar manner, obtaining 1454 images also organized into categories and with page titles.

\section{Machine Learning Results}

Two different kinds of models were tested on subsets of the data. A classifier was trained to test how visually distinguishable the given classes were and a generative model was trained to see if new instances of known illusions could be created by naively applying existing methods for image generation.

\subsection{Classifier Results}

\begin{figure}
    \centering
    \includegraphics[scale=0.4]{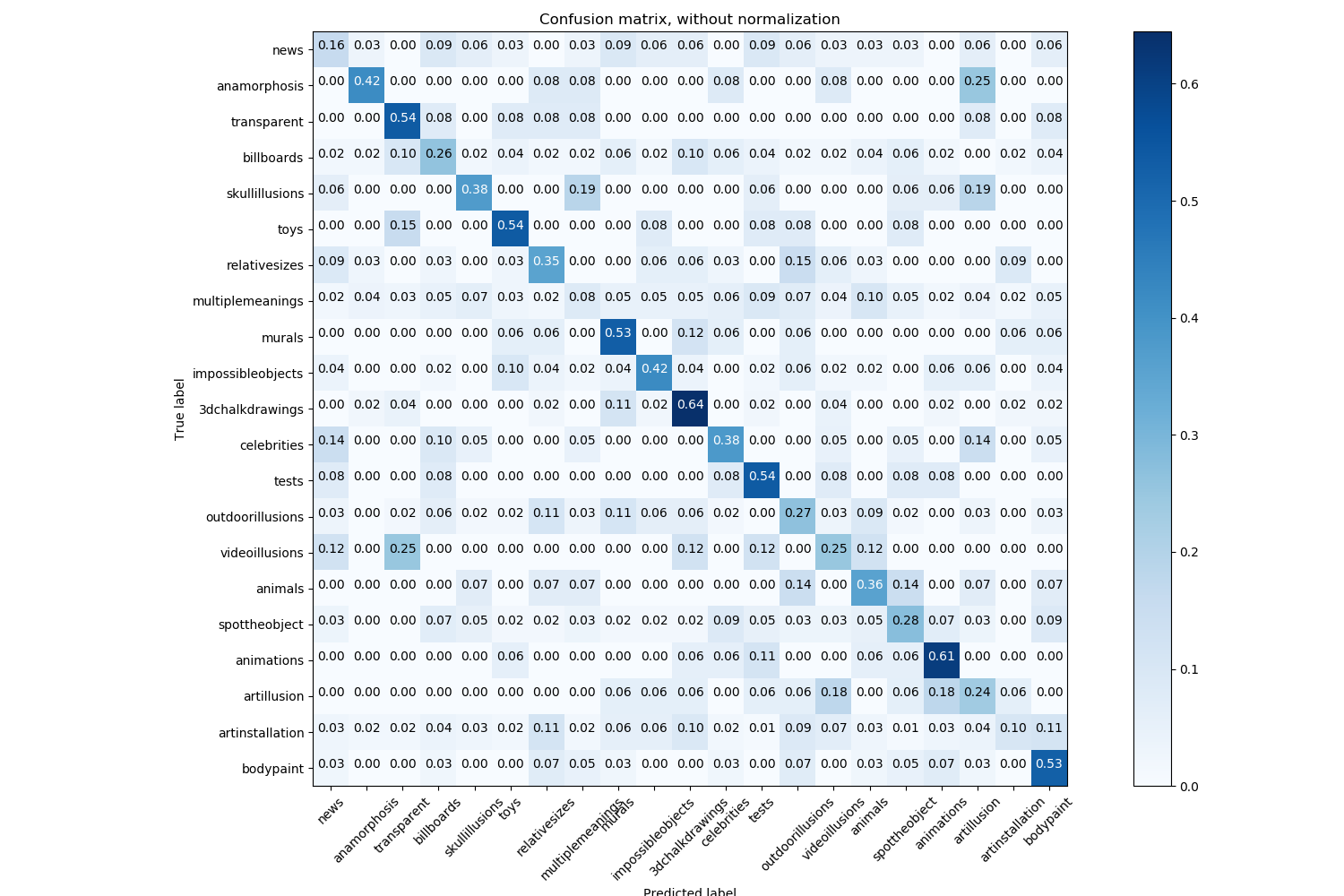}
    \caption{Confusion matrix for a classifier trained on the MoIllusions data.}
    \label{fig:confusion}
\end{figure}

A pretrained ``bottleneck" model \cite{BottleneckTurorial} was used to classify images from Mighty Optical Illusions. Only the last few layers had to be retrained, making use of transfer learning from a much larger dataset to learn to classify images in general. Each image in the training data may belong to multiple classes, which was not accounted for in the model. For the purpose of early dataset evaluation this flaw can be overlooked, but for more complete testing a multiclass model would need to be used. The results of training can be seen in Figure \ref{fig:confusion}.

\par

The model performed significantly better then random, meaning that the given classes are meaningful in a way the can be detected using a model trained on normal classes of images. A more in depth study could reveal more about how the neural network is able to distinguish these classes, such as the methods used in \cite{1710.00935}.

\subsection{Generative Adversarial Network Results}

\begin{figure}
    \centering
    \includegraphics[scale=0.4]{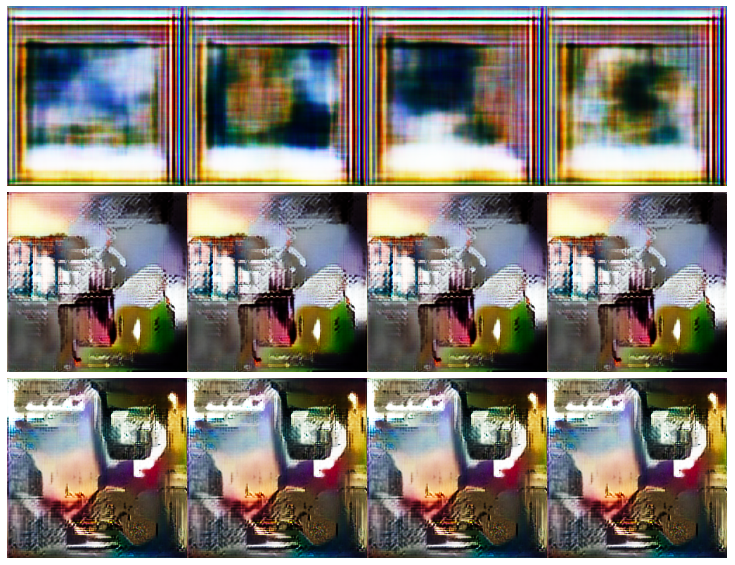}
    \caption{Failure of GAN to generate imagery similar to the dataset. Top to bottom is the progression from start to finish. Horizontal images are from the same training step but with different random input vectors \cite{1406.2661}. The lack of variety is abnormal and may lead to insights into how to correct the problem. }
    \label{fig:gan}
\end{figure}

A trial run using a generative adversarial network was attempted. Using HyperGAN \cite{HyperGAN} on a hand picked subset of the data with no hyperparameter optimization, nothing of value was created after 7 hours of training on an Nvidia Tesla K80. The training progression is shown in Figure \ref{fig:gan}. Possible improvements could be pretraining on a larger dataset, tweaking hyperparameters, and using dataset expansion techniques. 

\section{Future Work}

The only optical illusions known to humans have been created by evolution (for instance, eye patterns in butterfly wings) or by human artists. Both artistic designers of illusion images and the glacial process of evolution have access to active vision systems to verify their work against. An illusion artist can make an attempt at creating an illusion, observe its effect on their eyes, and add or remove elements to try to create a more powerful illusion. In an evolutionary process, every agent has a physical appearance and a vision system, allowing for patterns to be verified in their environment constantly. A GAN trained on existing illusions would have none of these advantages, and it seems unlikely that it could learn to trick human vision without being able to understand the principles behind these illusions.
Because of these limitations, it seems that a dataset of illusion images might not be sufficient to create new illusions and a deeper understanding of human vision would need to be obtained somehow. This could be by having a human giving feedback as the network learned, or by learning an accurate proxy for human vision and trying to deceive the proxy as in \cite{1802.08195}. 


\appendix

\section{Downloading the Dataset}
\label{appendix-sec1}
Images are currently hosted on the machine learning cloud platform ``Floydhub" and can be downloaded
without an account.

\begin{itemize}
\item \url{https://www.floydhub.com/robertmax/datasets/illusions-jpg}
\par
This contains all images that were downloaded, using the same numbering scheme as the metadata on the linked github repository.
\item \url{https://www.floydhub.com/robertmax/datasets/illusions-filtered}
\par
This folder contains images hand picked for having obvious visual effects without having to follow special instructions.

\end{itemize}


\bibliographystyle{elsarticle-num}

\bibliography{main}

\end{document}